\documentclass[runningheads]{llncs}
\usepackage{graphicx}

\usepackage{tikz}
\usepackage{comment}
\usepackage{amsmath,amssymb} 
\usepackage{color}

\usepackage[accsupp]{axessibility}  

\usepackage[width=122mm,left=12mm,paperwidth=146mm,height=193mm,top=12mm,paperheight=217mm]{geometry}

\usepackage{multirow}
\usepackage{graphicx}
\usepackage{amsmath}
\usepackage{makecell}
\usepackage{float}
\usepackage{soul}
\usepackage{arydshln}
\usepackage[pdftex, pagebackref=true,breaklinks=true,letterpaper=true,colorlinks,bookmarks=false, citecolor=green, linkcolor=red]{hyperref}
\usepackage{breakcites}

\begin{document}
\pagestyle{headings}
\mainmatter
\def\ECCVSubNumber{339}  


\title{Efficient Long-Range Attention Network for Image Super-resolution}

\titlerunning{Efficient Long-Range Attention Network for Image Super-resolution}
%
\author{
    Xindong Zhang \inst{1,2,}\thanks{Equal contribution} \and
    Hui Zeng  \inst{2,\star}  \and
    Shi Guo   \inst{1}        \and 
    Lei Zhang \inst{1,}\thanks{Corresponding author}
}
\authorrunning{Xindong Zhang et al.}
%
\institute{
    Dept. of Computing, The Hong Kong Polytechnic University \and 
    OPPO Research, China \\
    \email{\{csxdzhang, csshiguo, cslzhang\}@comp.polyu.edu.hk, cshzeng@gmail.com} 
}


\maketitle

\begin{abstract}

Recently, transformer-based methods have demonstrated impressive results in various vision tasks, including image super-resolution (SR), by exploiting the self-attention (SA) for feature extraction. However, the computation of SA in most existing transformer based models is very expensive, while some employed operations may be redundant for the SR task. This limits the range of SA computation and consequently the SR performance. In this work, we propose an efficient long-range attention network (ELAN) for image SR. Specifically, we first employ shift convolution (shift-conv) to effectively extract the image local structural information while maintaining the same level of complexity as 1x1 convolution, then propose a group-wise multi-scale self-attention (GMSA) module, which calculates SA on non-overlapped groups of features using different window sizes to exploit the long-range image dependency. A highly efficient long-range attention block (ELAB) is then built by simply cascading two shift-conv with a GMSA module, which is further accelerated by using a shared attention mechanism. Without bells and whistles, our ELAN follows a fairly simple design by sequentially cascading the ELABs. Extensive experiments demonstrate that ELAN obtains even better results against the transformer-based SR models but with significantly less complexity. The source code can be found at \url{https://github.com/xindongzhang/ELAN}.



\end{abstract}

\section{Introduction}
Singe image super-resolution (SR) aims at reproducing a high-resolution (HR) output from its degraded low-resolution (LR) counterpart. In recent years, deep convolutional neural network (CNN) \cite{lecun2015deep} based SR models \cite{dong2015image,dong2016accelerating,shi2016real,lim2017enhanced,caballero2017real,tao2017detail,zhang2018image,sajjadi2018frame,wang2019edvr} have become prevalent for their strong capability in recovering or generating \cite{ledig2017photo,wang2018esrgan} image high-frequency details, showing high practical value in image and video restoration, transition and display. 
However, many CNN-based methods extract local features with spatially invariant kernels, which are inflexible to adaptively model the relations among pixels. In addition, to enlarge the receptive field, the CNN-based SR models tend to employ very deep and complicated network topology \cite{zhang2018image,dai2019second,zhou2020cross,niu2020single,mei2021image} to recover more details, resulting in much computational resource consumption.

Recently, the transformer based methods have shown impressive performance on natural language processing tasks \cite{brown2020language,fedus2021switch,liu2019roberta,radford2018improving} for their strong ability on modeling the self-attention (SA) of input data. The great success of transformers on NLP has inspired researchers to explore their application to computer vision tasks, and interesting results have been shown on several high-level vision tasks \cite{dosovitskiy2020image,li2021localvit,liu2021transformer,liu2021swin,ramachandran2019stand,touvron2021training}. Generally speaking, the input image is first divided into non-overlapped patches with different scales as tokens, and then the local feature extraction and SA modules are applied to the collection of patches. Though SA has proven to be effective for high-level vision tasks, its complexity grows quadratically with the input feature size, limiting its utilization in low-level vision tasks such as SR, where the feature size is usually very large. 

A few attempts have been made to reduce the computational cost of SA in the application of SR. Mei \emph{et al}. \cite{mei2021image} divided an image into non-overlapped patches for modeling local feature and SA independently, which however may introduce border effect and deteriorate the visual quality of restored images. SwinIR \cite{liang2021swinir} follows the design of Swin Transformer \cite{liu2021swin}, where the local feature is first extracted by two cascaded $1\times1$ convolutions and then SA is calculated within a small sized window (i.e., $8\times8$) with a shifting mechanism to build connections with other windows. However, local features extracted by $1\times1$ convolutions with small receptive field may produce weak feature representations to calculate the SA for long-range modeling. Furthermore, calculating SA with a small window size restricts the ability of modeling long-range dependency among image pixels. Restormer \cite{zamir2021restormer} calculates the SA with dependency on the channel space which remains applicable to large images. However, modeling dependency on channel space may sacrifice some useful spatial information of textures and structures which is important for reproducing high quality HR images.

In this paper, we aim to develop an effective and efficient way to exploit image long-range attention for image SR with a simple network architecture. The existing transformer-based models such as SwinIR \cite{liang2021swinir} have many redundant and fragmented components which are not cost-effective for the SR task, such as relative position bias, masking mechanism, layer normalization, and several sub-branches created with residual shortcut. We therefore aim to build a highly neat and efficient SR model, where the LR to HR image mapping is simply built by stacking local feature extraction operations and SA sequentially. Two successive shift convolution (shift-conv) operations are used to efficiently extract local structural information, while a shift-conv has larger receptive field but shares the same arithmetic complexity as a $1\times1$ convolution. For SA calculation, we propose a group-wise multi-scale self-attention (GMSA) operator, which divides features into different groups of different window sizes and calculates SA separately. This strategy provides a larger receptive field than a small fixed-size window for long-range attention modeling, while being more flexible than a large fixed-size window for reducing the computation resource cost. Furthermore, a shared attention mechanism is proposed to accelerate the calculation for successive GMSA modules. Arming with the shift-conv and GMSA with shared attention, an efficient long-range attention network, namely ELAN, is readily obtained for SR. Our contributions are summarized as follows:


\begin{itemize}
  \item [1)] We propose an efficient long-range attention block to model image long-range dependency, which is important for improving the SR performance.
  \item [2)] We present a fairly simple yet powerful network, namely ELAN, which records new state-of-the arts for image SR with significantly less complexity than existing vision transformer based SR methods.
\end{itemize}

\section{Related work}
Numerous deep learning based image SR methods have been developed in the past decade. Here we briefly discuss the related work from the perspective of CNN-based methods and transformer-based methods.  

\subsection{CNN-based SR methods}

CNN-based methods have demonstrated impressive performance in the SR task. SRCNN \cite{dong2015image} makes the first attempt to employ CNN for image SR by learning a non-linear mapping from the bicubically upsampled LR image to the HR output with only three convolution layers. Kim \emph{et al}. \cite{kim2016accurate} deepened the network with VGG-19 and residual learning and achieved much better performance. Since the pre-upsampling strategy increases the amount of input data to CNN and causes large computational cost, FSRCNN \cite{dong2016accelerating} adopts a post-upsampling strategy to accelerate the CNN model. An enhanced residual block was proposed in \cite{lim2017enhanced} to train the deep model without batch normalization, and the developed EDSR network won the first prize of the NTIRE2017 challenge \cite{agustsson2017ntire}. 

To build more effective models for SR, the recently developed methods tend to employ deeper and more complicated architectures as well as the attention techniques. Zhang \emph{et al}. proposed a residual-in-residual structure coupled with channel attention to train a very deep network over 400 layers. Other works like MemNet \cite{tai2017memnet} and RDN \cite{zhang2018residual} are designed by employing the dense blocks \cite{huang2017densely} to utilize the intermediate features from all layers.  In addition to increasing the depth of network, some other works,
such as SAN \cite{dai2019second}, NLRN \cite{liu2018non}, HAN \cite{niu2020single} and NLSA \cite{mei2021image}, excavate the feature correlations along the spatial or channel dimension to boost the SR performance. Our proposed ELAN takes the advantage of fast local feature extraction, while models the long-range dependency of features via efficient group-wise multi-scale self-attention.

\subsection{Transformer-based SR methods}
The breakthrough of transformer networks in natural language processing (NLP) inspired of use of self-attention (SA) in computer vision tasks. The SA mechanism in transformers can effectively model the dependency across data, and it has achieved impressive results on several high-level vision tasks, such as image classification \cite{dosovitskiy2020image,li2021localvit,liu2021transformer,liu2021swin,ramachandran2019stand,touvron2021training}, image detection \cite{carion2020end,liu2020deep,liu2021swin,touvron2021training}, and segmentation \cite{cao2021swin,liu2021swin,wu2020visual,zheng2021rethinking}. Very recently, transformer has also been applied to low-level vision tasks \cite{liang2021swinir,wang2021uformer,zamir2021restormer,chen2021pre}. IPT \cite{chen2021pre} is an extremely large pre-trained model for various low-level vision tasks based on the standard vision transformer. It computes both the local feature and SA on non-overlapped patches, which however may lose some useful information for reproducing image details. SwinIR \cite{liang2021swinir} hence adapts the Swin Transformer \cite{liu2021swin} to image restoration, which combines the advantages of both CNNs and transformers. Though SwinIR has achieved impressive results for image SR, its network structure is mostly borrowed from the Swin Transformer, which is designed for high-level vision tasks. In particular, the network design of SwinIR is redundant for the SR problem, and it calculates SA on small fixed-size windows, preventing it from exploiting long-range feature dependency. Our proposed ELAN is not only much more efficient than SwinIR but also able to compute the SA in larger windows. 



\section{Methodology}
In this section we first present the pipeline of our efficient long-range attention network (ELAN) for SR tasks, and then discuss in detail its key component, the efficient long-range attention block (ELAB).

\subsection{Overall Pipeline of ELAN}

\begin{figure}[t]
  \centering
  \includegraphics[width=1.0\linewidth]{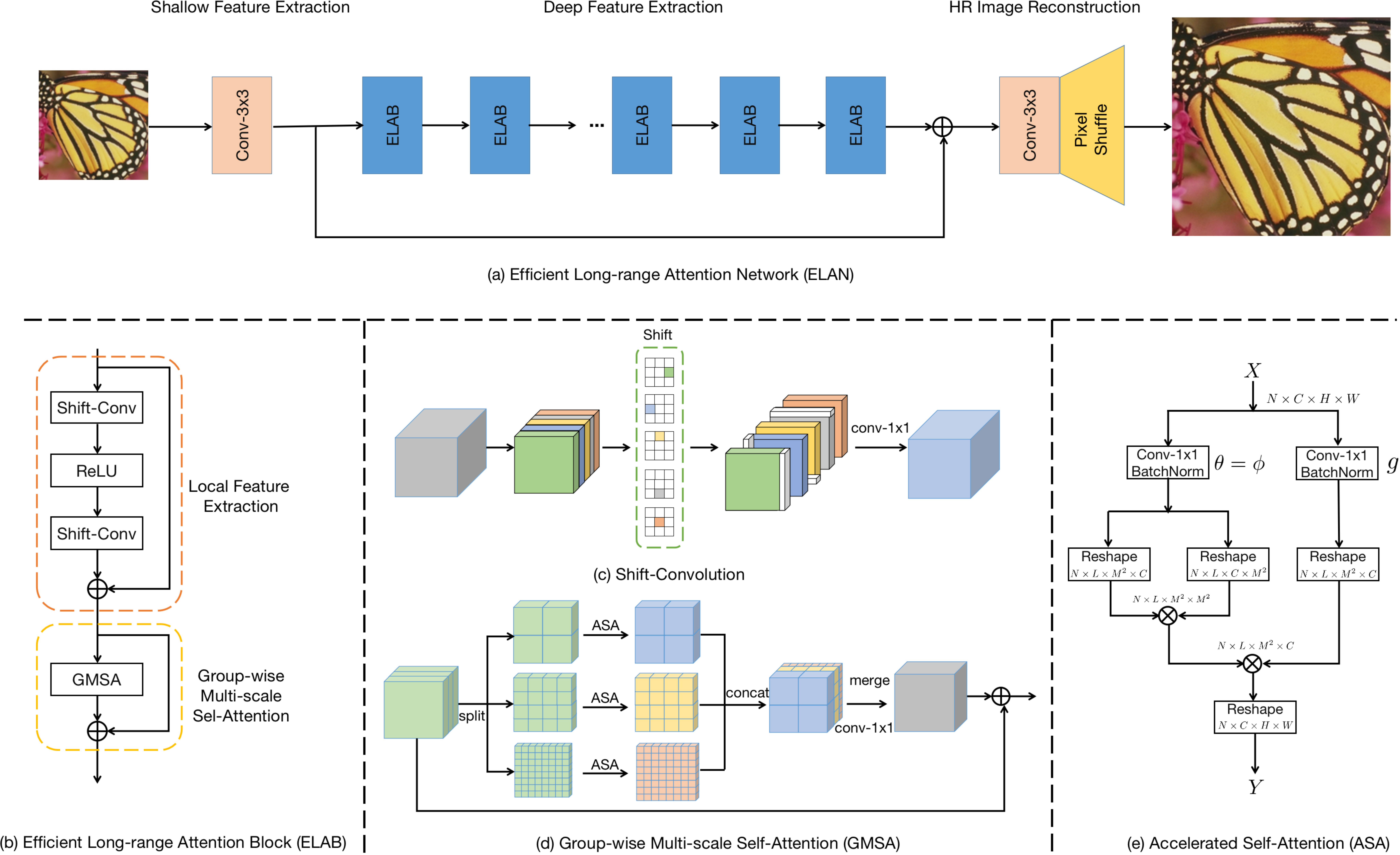}
  \caption{Illustration of the proposed efficient long-range attention network (ELAN). (a) The overall pipeline of ELAN, which contains several ELABs, two $3\times3$ convolutions and one pixel shuffle operator. (b) The architecture of the efficient long-range attention block (ELAB). (c) Illustration of shift-conv, which is composed of a shift operation followed by one $1\times1$ convolution. (d) Illustration of the computation of group-wise multi-scale self-attention (GMSA). (e) Illustration of our accelerated self-attention (ASA) computation.}
  \label{ELAN_figure}
\end{figure}

The overall pipeline of ELAN is shown in Figure \ref{ELAN_figure}(a), which consists of three modules: shallow feature extraction, ELAB based deep feature extraction, and HR image reconstruction. The network follows a fairly simple topology with a global shortcut connection from the shallow feature extraction module to the output of deep feature extraction module before fed into the HR reconstruction module. In specific, given a degraded LR image $X_{l} \in \mathbb{R}^{3 \times H \times W}$, where $H$ and $W$ are the height and width of the LR image, respectively, we first apply the shallow feature extraction module, denoted by $H_{SF}(\cdot)$, which consists of only a single $3\times3$ convolution, to extract the local feature $X_s \in \mathbb{R}^{C \times H \times W}$:
\begin{equation}
    X_s = H_{SF}(X_l)
\label{shadow_feature_extraction}
\end{equation}
where $C$ is the channel number of the intermediate feature. 

$X_s$ then goes to the deep feature extraction module, denoted by $H_{DF}(\cdot)$, which is composed of $M$ cascaded ELABs. That is:
\begin{equation}
    X_d = H_{DF}(X_s),
\label{deep_feature_extraction}
\end{equation}
where $X_d \in \mathbb{R}^{C\times H \times W}$ denotes the output. By taking $X_d$ and $X_s$ as inputs, the HR image $X_h$ is reconstructed as:
\begin{equation}
    X_h = H_{RC}(X_s + X_d),
\label{reconstruction}
\end{equation}
where $H_{RC}$ is the reconstruction module. There are some choices for the design of the reconstruction module \cite{dong2015image,dong2016accelerating,shi2016real,lim2017enhanced}. To achieve high efficiency, we build it simply with a single $3\times3$ convolution and a pixel shuffle operation. 

The ELAN can be optimized with the commonly used loss functions for SR, such as $L_2$ \cite{dong2015image,kim2016accurate,tai2017image,tai2017memnet}, $L_1$ \cite{lai2017deep,lim2017enhanced,zhang2018residual} and perceptual losses \cite{huang2017densely,sajjadi2017enhancenet}. For simplicity, given a number of $N$ ground-truth HR images $\{X_{t,i}\}_{i=1}^{N}$, we optimize the parameters of ELAN by minimizing the pixel-wise $L_1$ loss:
\begin{equation}
    \mathcal{L}= \frac{1}{N} \sum_{i=1}^{N} || X_{h,i} - X_{t,i} ||_1.
\label{loss_function}
\end{equation}
\noindent
The Adam optimizer \cite{kingma2014adam} is employed to optimize our ELAN for its good performance in low-level vision tasks. 

\subsection{Efficient Long-range Attention Block (ELAB)}

As shown in Figure \ref{ELAN_figure}(b), our ELAB is composed of a local feature extraction module and a group-wise multi-scale attention (GMSA) module, both equipped with the residual learning strategy. 

\textbf{Local feature extraction}. Given the intermediate features $X$, previous researches mostly extract the local features via multi-layer perception or two cascaded $1\times1$ convolutions, which however have only $1\times1$ receptive field. To enlarge the receptive field for more effective local feature extraction, we utilize two shift-conv \cite{wu2018shift} with a simple ReLU activation between them. As shown in Figure \ref{ELAN_figure}(c), the shift-conv is composed of a set of shift operations and one $1\times1$ convolution. Specifically, we split the input feature equally into five groups, and move the first four groups of feature along different spatial dimensions, including left, right, top, bottom, while the last group remains unchanged. Therefore, the followed $1\times1$ convolution can leverage information from neighboring pixels. Without introducing additional learnable parameters and much computation, shift-conv can provide larger receptive fields while maintaining almost the same arithmetic complexity as $1\times1$ convolution.


\textbf{Group-wise multi-scale self-attention (GMSA).} Given a feature map of $C\times H\times W$, the computational complexity of the window-based self-attention \cite{liu2021swin,liang2021swinir} using $M\times M$ non-overlapped windows is $2M^2HWC$. The window size $M$ determines the range of SA calculation, and a larger $M$ contributes to exploit more self-similarity information. However, directly enlarging $M$ will quadratically increase the computational cost and resources. To more efficiently calculate the long-range SA, we propose the GMSA module, which is illustrated in Figure \ref{ELAN_figure}(d). We first split the input feature  $X$ into $K$ groups, denoted by $\{X_k\}_{k=1}^{K}$, then calculate SA on the $k$-th group of features using window size $M_k$. In this way, we can flexibly control the computational cost by setting the ratio of different window size. For example, supposing the $K$ groups of features are equally split with $\frac{C}{K}$ channels, the computational cost of $K$ groups of SA is $\frac{2}{K}(\sum_kM_k^2)HWC$. The SA calculated on different groups are then concatenated and merged via a $1\times1$ convolution. 

\textbf{Accelerated self-attention (ASA).} The calculation of SA is computation and memory-intensive in existing transformer models \cite{cao2021swin,liu2021swin,liang2021swinir,wang2021uformer}. We make several modifications to accelerate the calculation of SA, especially in the inference stage. First, we discard the layer normalization (LN), which is widely employed in previous transformer models \cite{cao2021swin,liu2021swin,liang2021swinir,wang2021uformer}, because the LN fragments the calculation of SA into many element-wise operations, which are not friendly for efficient inference. Instead, we utilize batch normalization (BN) \cite{ioffe2015batch} to stabilize the training process. It is worth mentioning that the BN can be merged into the convolution operation, which does not cause additional computation cost in the inference stage. Second, the SA in SwinIR \cite{liang2021swinir} is calculated on the embedded Gaussian space, where three independent $1\times1$ convolutions, denoted by $\theta$, $\phi$ and $g$, are employed to map the input feature $X$ into three different feature maps. We set $\theta=\phi$ and calculate the SA in the symmetric embedded Gaussian space \cite{liu2018non,buades2005non}, which can save one $1\times1$ convolution in each SA. This modification further alleviates the computation and memory cost of SA without sacrificing the SR performance. Our ASA is shown in Figure \ref{ELAN_figure}(e).

\begin{figure}[t]
  \centering
  \includegraphics[width=1.0\linewidth]{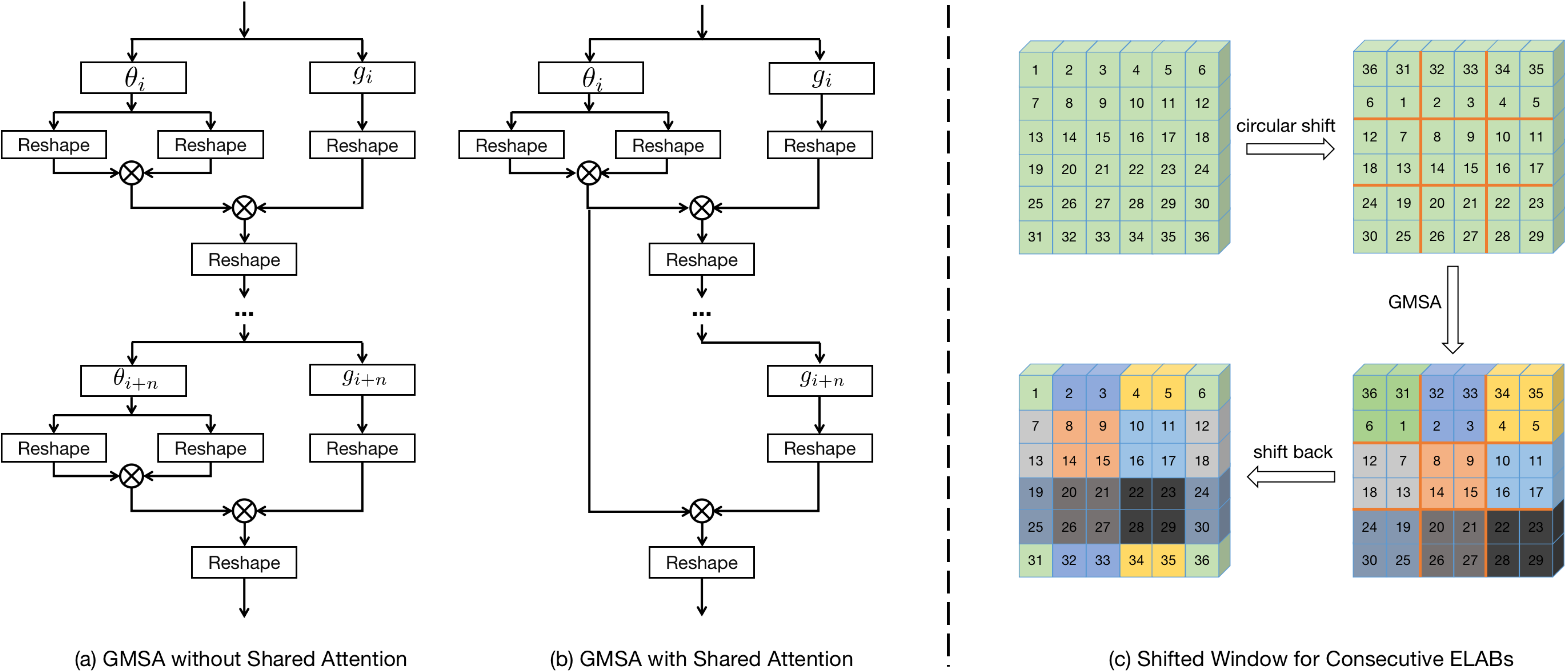}
  \caption{(a) and (b) illustrate the calculation of GMSA without and with the shared attention mechanism, respectively. (c) Illustration of the shifted window mechanism used in consecutive ELABs. }
  \label{ELAB_mechanism_figure}
\end{figure}

\textbf{Shared attention.} Despite the above acceleration, one single forward pass of SA still consists of two $1\times1$ convolutions and four IO-intensive reshape operations. Although the reshape operation is FLOPs-free, it is time-consuming due to large feature size in the SR task. To further accelerate the SA computation of the entire network, we propose to share the attention scores among adjacent SA modules. As shown in Figure \ref{ELAB_mechanism_figure}(b), the calculated attention scores in the $i$-th SA module is directly re-used by the following $n$ SA modules on the same scale.
In this way, we can avoid $2n$ reshape and $n$ $1\times1$ convolution operations for the following $n$ SAs. We found that the proposed shared attention mechanism only leads to slight drop in SR performance by using a a small number of $n$ (e.g., 1 or 2), while it saves much computation resources during inference.

\textbf{Shifted window.} The calculated SA on the group-wise multi-scale windows still lacks connection across local windows within the same scale. We improve the shifted window mechanism of SwinIR \cite{liang2021swinir} to reach a simple yet effective shifting scheme for the SR task. The whole process is visualized in Figure \ref{ELAB_mechanism_figure}(c). We first employ circular shift to the feature along the diagonal direction and calculate GMSA on the shifted feature. Then we shift the result back via inverse circular shift. The circular shift with half window size leads to a new partition of feature map and introduces connections among neighboring non-overlapped windows in the previous GMSA module. Although some pixels on the border are shifted to distant areas via circular shift, we found it has negligible impact on the final SR performance since such pixels only occupy a small portion of the entire feature map in the SR task. Benefiting from the circular shift mechanism, we remove the masking strategy and relative positional encoding adopted in SwinIR \cite{liang2021swinir}, making our network neater and more efficient.


\section{Experiments}

In this section, we conduct extensive experiments to quantitatively and qualitatively validate the superior performance of our ELAN for light-weight and classic SR tasks on five SR benchmark datasets. We also present comprehensive ablation studies to evaluate the design of our proposed ELAN.

\begin{table}[t]
\caption{Performance comparison of different light-weight SR models on five benchmarks. PSNR/SSIM on Y channel are reported on each dataset. \#Params and \#FLOPs are the total number of network parameters and floating-point operations, respectively. All the efficiency proxies (\#Params, \#FLOPs and Latency) are measured under the setting of upscaling SR images to $1280\times720$ resolution on all scales. Best and second best PSNR/SSIM indexes are marked in \textcolor{red}{red} and \textcolor{blue}{blue} colors, respectively. The CNN-based methods and transformer-based methods are separated via a dash line for each scaling factor. '--' means that the result is not available.}
\label{LightWeightBenchmarkResults}
\resizebox{\textwidth}{45mm}{
    \begin{tabular}{c|c|c|c|c|c|c|c|c|c}
    \hline
    Scale & Model & \makecell{\#Params \\ (K)} & \makecell{\#FLOPs \\ (G)} & \makecell{Latency \\ (ms)} & \makecell{Set5 \cite{bevilacqua2012low}\\PSNR/SSIM} & \makecell{Set14 \cite{zeyde2010single}\\PSNR/SSIM} & \makecell{B100\cite{martin2001database}\\PSNR/SSIM} & \makecell{Urban100 \cite{huang2015single}\\PSNR/SSIM} & \makecell{Manga109 \cite{matsui2017sketch}\\PSNR/SSIM} \\
    \hline
    \multirow{8}{*}{$\times$ 2}
        & CARN \cite{ahn2018fast} & 1,592 & 222.8 & 72 & 37.76/0.9590 & 33.52/0.9166 & 32.09/0.8978 & 31.92/0.9256 & 38.36/0.9765\\
        & EDSR-baseline \cite{lim2017enhanced} & 1370 & 316.3 & 71 & 37.99/0.9604 & 33.57/0.9175 & 32.16/0.8994 & 31.98/0.9272 & 38.54/0.9769 \\
        & IMDN \cite{hui2019lightweight} & 694 & 158.8 & 54 & 38.00/0.9605 & 33.63/0.9177 & 32.19/0.8996 & 32.17/0.9283 & 38.88/0.9774\\
        & LAPAR-A \cite{li2020lapar} & 548 & 171.0 & 73 & 38.01/0.9605 & 33.62/0.9183 & 32.19/0.8999 & 32.10/0.9283 & 38.67/0.9772\\
        & LatticeNet \cite{luo2020latticenet} & 756 & 169.5 & 66 & 38.06/0.9607 & 33.70/0.9187 & 32.20/0.8999 & 32.25/0.9288 & ---/---\\
        \cdashline{2-10}[5pt/5pt]
        & ESRT \cite{lu2021efficient} & 677 & --- & --- & 38.03/0.9600 & 33.75/0.9184 & 32.25/0.9001 & 32.58/0.9318 & \textcolor{red}{39.12}/0.9774 \\
        & SwinIR-light \cite{liang2021swinir} & 878 & 195.6 & 1007 & \textcolor{blue}{38.14}/\textcolor{red}{0.9611} & \textcolor{blue}{33.86/0.9206} & \textcolor{red}{32.31/0.9012} & \textcolor{red}{32.76/0.9340} & \textcolor{red}{39.12/0.9783} \\
        & ELAN-light (ours)& 582 & 168.4 & 230 & \textcolor{red}{38.17/0.9611} & \textcolor{red}{33.94/0.9207} & \textcolor{blue}{32.30}/\textcolor{red}{0.9012} & \textcolor{red}{32.76/0.9340} & 39.11/\textcolor{blue}{0.9782}\\
    \hline
    \hline
    \multirow{8}{*}{$\times$ 3}
        & CARN \cite{ahn2018fast} & 1,592 & 118.8 & 39 & 34.29/0.9255 & 30.29/0.8407 & 29.06/0.8034 & 28.06/0.8493 & 33.50/0.9440 \\
        & EDSR-baseline \cite{lim2017enhanced} & 1555 & 160.2 & 37 & 34.37/0.9270 & 30.28/0.8417 & 29.09/0.8052 & 28.15/0.8527 & 33.45/0.9439 \\
        & IMDN \cite{hui2019lightweight} & 703 & 71.5 & 27 & 34.36/0.9270 &  30.32/0.8417 & 29.09/0.8046 & 28.17/0.8519 & 33.61/0.9445\\
        & LAPAR-A \cite{li2020lapar} & 544 & 114.0 & 55 & 34.36/0.9267 & 30.34/0.8421 & 29.11/0.8054 & 28.15/0.8523 & 33.51/0.9441\\
        & LatticeNet \cite{luo2020latticenet} & 765 & 76.3 & 33 & 34.40/0.9272 & 30.32/0.8416 & 29.10/0.8049 & 28.19/0.8513 & ---/---\\
        \cdashline{2-10}[5pt/5pt]
        & ESRT \cite{lu2021efficient} & 770 & --- & --- & 34.42/0.9268 & 30.43/0.8433 & 29.15/0.8063 & 28.46/0.8574 & 33.95/0.9455 \\
        & SwinIR-light \cite{liang2021swinir} & 886 & 87.2 & 445 & \textcolor{red}{34.62/0.9289} & \textcolor{blue}{30.54}/\textcolor{red}{0.8463} & \textcolor{blue}{29.20}/\textcolor{red}{0.8082} & \textcolor{blue}{28.66}/\textcolor{red}{0.8624} & \textcolor{blue}{33.98}/\textcolor{red}{0.9478}\\
        & ELAN-light (ours)& 590 & 75.7 & 105 & \textcolor{blue}{34.61/0.9288} & \textcolor{red}{30.55/0.8463} & \textcolor{red}{29.21}/\textcolor{blue}{0.8081} & \textcolor{red}{28.69/0.8624} & \textcolor{red}{34.00/0.9478}\\
    \hline
    \hline
    \multirow{8}{*}{$\times$ 4}
        & CARN \cite{ahn2018fast} & 1,592 & 90.9 & 30 & 32.13/0.8937 & 28.60/0.7806 & 27.58/0.7349 & 26.07/0.7837 & 30.47/0.9084\\
        & EDSR-baseline \cite{lim2017enhanced} & 1518 & 114.0 & 28 & 32.09/0.8938 & 28.58/0.7813 & 27.57/0.7357 & 26.04/0.7849 & 30.35/0.9067 \\
        & IMDN \cite{hui2019lightweight} & 715 & 40.9 & 19 & 32.21/0.8948 & 28.58/0.7811 & 27.56/0.7353 & 26.04/0.7838 & 30.45/0.9075\\
        & LAPAR-A \cite{li2020lapar} & 659 & 94.0 & 47 & 32.15/0.8944 & 28.61/0.7818 & 27.61/0.7366 & 26.14/0.7871 & 30.42/0.9074\\
        & LatticeNet \cite{luo2020latticenet} & 777 & 43.6 & 23 & 32.18/0.8943 & 28.61/0.7812 & 27.57/0.7355 & 26.14/0.7844 & ---/--- \\
        \cdashline{2-10}[5pt/5pt]
        & ESRT \cite{lu2021efficient} & 751 & --- & --- & 32.19/0.8947 & 28.69/0.7833 & 27.69/0.7379 & 26.39/0.7962 & 30.75/0.9100 \\
        & SwinIR-light \cite{liang2021swinir} & 897 & 49.6 & 271 & \textcolor{red}{32.44/0.8976} & \textcolor{blue}{28.77}/\textcolor{red}{0.7858} & \textcolor{red}{27.69/0.7406} & \textcolor{blue}{26.47/0.7980} & \textcolor{red}{30.92/0.9151}\\
        & ELAN-light (ours) & 601 & 43.2 & 62 & \textcolor{blue}{32.43/0.8975} & \textcolor{red}{28.78/0.7858} & \textcolor{red}{27.69/0.7406} & \textcolor{red}{26.54/0.7982} & \textcolor{red}{30.92}/\textcolor{blue}{0.9150}\\
    \hline
    \end{tabular}
}
\end{table}

\subsection{Experimental setup}

\textbf{Dataset and evaluation metrics.} We employ the DIV2K dataset \cite{timofte2017ntire} with 800 training images to train our ELAN model, and use the five benchmark datasets, including Set5\cite{bevilacqua2012low}, Set14\cite{zeyde2010single}, BSD100\cite{martin2001database}, Urban100\cite{huang2015single} and Manga109\cite{matsui2017sketch}, for performance comparison. PSNR and SSIM are used as the evaluation metrics, which are calculated on the Y channel after converting RGB to YCbCr format.
For efficiency comparison, we report the latency evaluated on a single NVIDIA 2080Ti GPU in the inference stage. We also report the number of network parameters and FLOPs as reference, although they may not be able to faithfully reflect the network complexity and efficiency. 

Note that since some competing methods do not release the source codes, we can only copy their PSNR/SSIM results from the original papers, but cannot report their results of latency and FLOPS.

\textbf{Training details}. Following SwinIR \cite{liang2021swinir}, we train two versions of ELAN with different complexity. The light-weight version, i.e., ELAN-light consists of 24 ELABs with 60 channels, while the normal version of ELAN has 36 ELABs with 180 channels. We calculate GMSA on three equally split scales with window size: $4\times4$, $8\times8$ and $16\times16$. By default, we set $n=1$ for the shared attention mechanism. Bicubic downsampling is used to generate training image pairs. We randomly crop 64 patches of size $64\times64$, and 32 patches of size $48\times48$ from the LR images as training mini-batch for the light-weight and normal ELAN models, respectively. We employ randomly rotating $90^{\circ}$, $180^{\circ}$, $270^{\circ}$ and horizontal flip for data augmentation. Both models are trained using the ADAM optimizer with $\beta_1=0.9$, $\beta_2=0.999$, and $\epsilon=10^{-8}$ for 500 epochs. The learning rate is initialized as $2\times10^{-4}$ and multiplied with $0.5$ after $\{250, 400, 425, 450, 475\}$-th epoch. The model training is conducted by Pytorch \cite{paszke2019pytorch} on NVIDIA 2080Ti GPUs.





\begin{figure}[!t]
  \centering
  \includegraphics[width=1.0\linewidth]{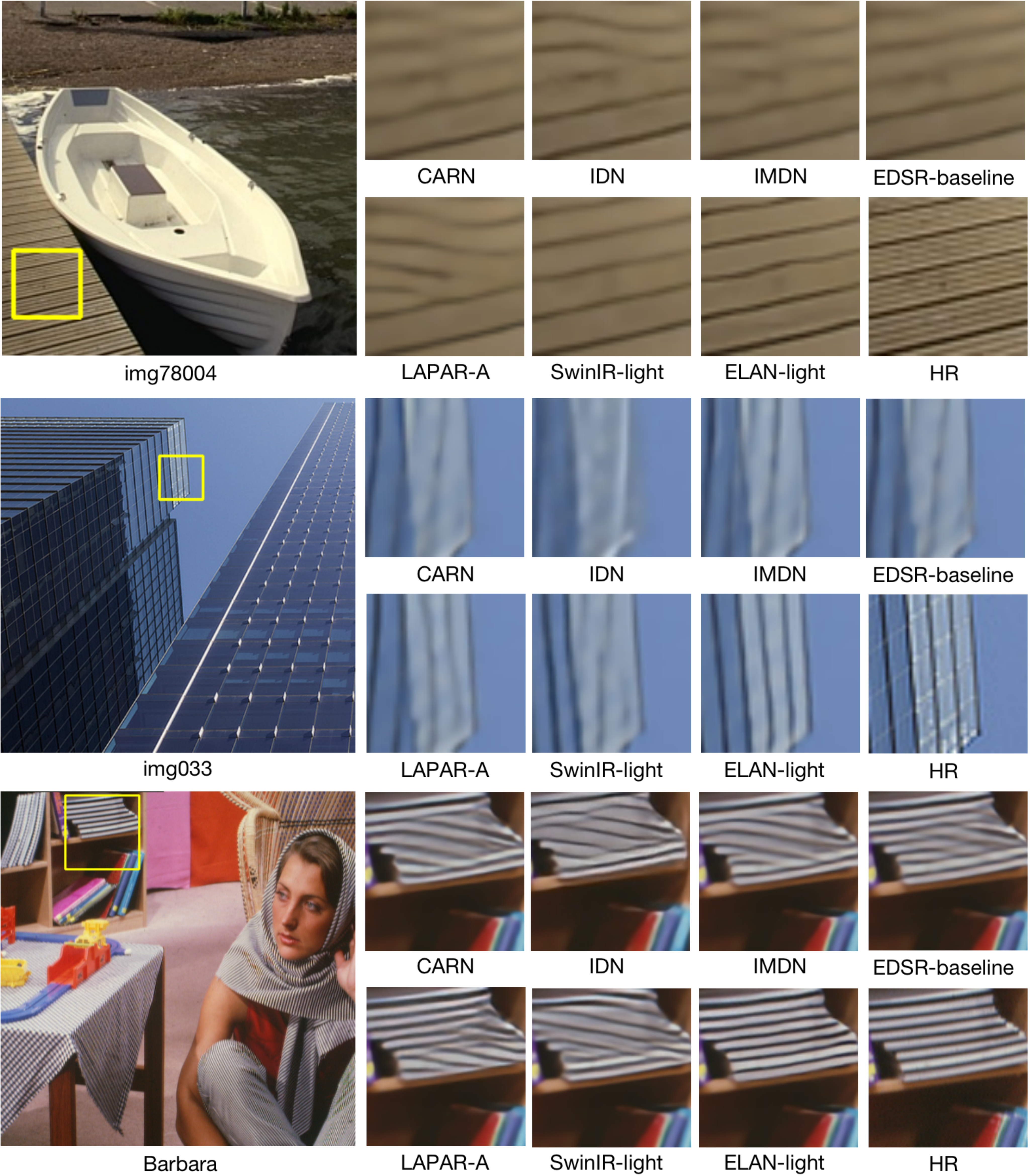}
  \caption{Qualitative comparison of state-of-the-art light-weight SR models for $\times4$ upscaling. ELAN-light can restore more accurate and clear structures than other models.
  }
  \label{VISUAL_ELAN_LIGHT_figure}
\end{figure}

\begin{figure}[!t]
  \centering
  \includegraphics[width=1.0\linewidth]{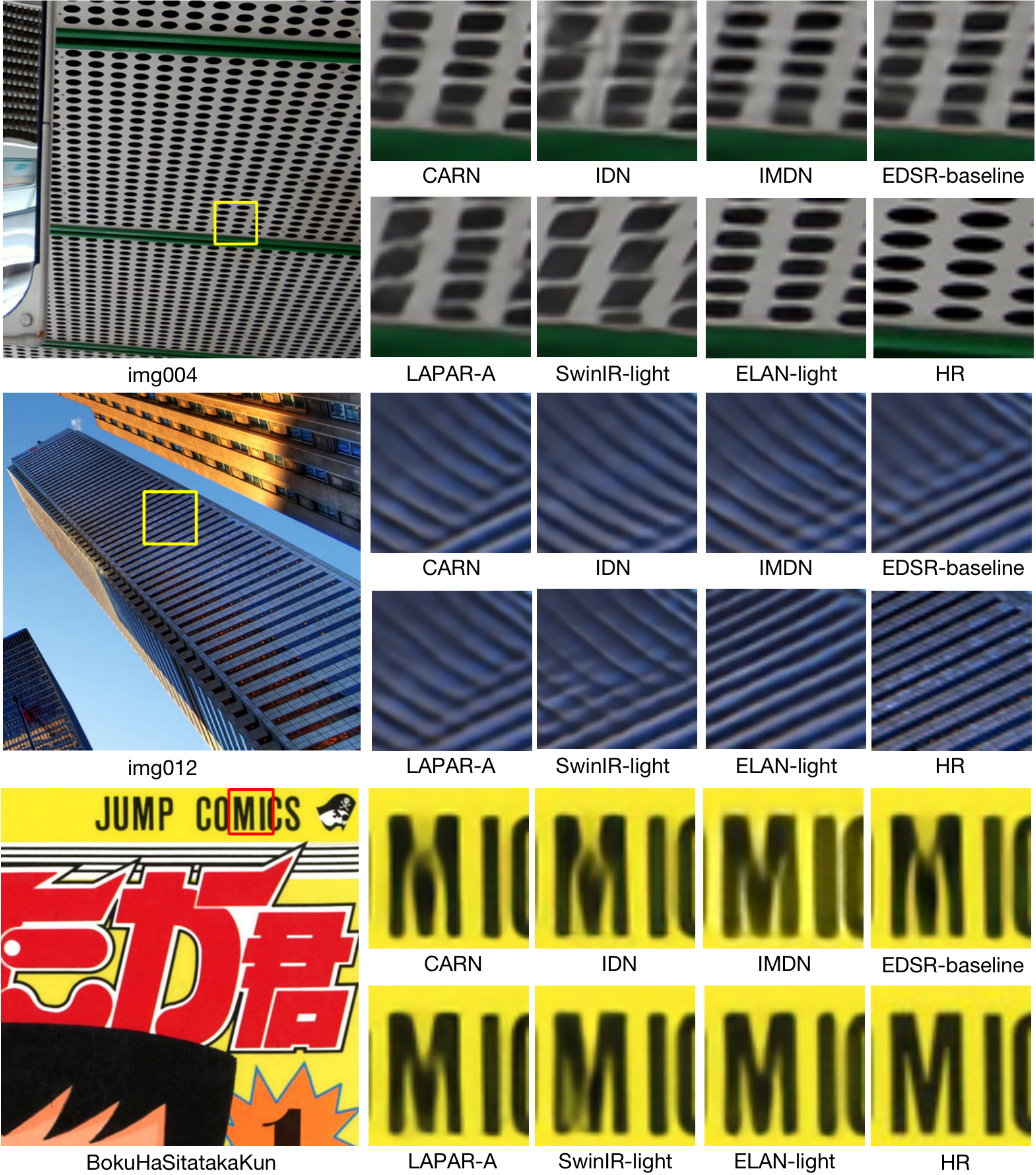}
  \caption{Qualitative comparison of state-of-the-art light-weight SR models for $\times4$ upscaling. ELAN-light can restore more accurate and clear structures than other models.
  }
  \label{VISUAL_ELAN_LIGHT_figure_extra}
\end{figure}

\subsection{Comparison with light-weight SR models}

We first compare our ELAN-light with state-of-the-art light-weight SR models, including CNN-based models CARN\cite{ahn2018fast}, IMDN \cite{hui2019lightweight}, LAPAR-A\cite{li2020lapar}, LatticeNet\cite{luo2020latticenet}, and transformer-based models ESRT \cite{lu2021efficient} and SwinIR-light\cite{liang2021swinir}. 

\textbf{Quantitative comparison}. The quantitative indexes of different methods are reported in Table \ref{LightWeightBenchmarkResults}. Several observations can be made from the table. First, with similar \#Params and \#FLOPs, the transformer-based methods especially SwinIR-light outperform much CNN-based methods on PSNR/SSIM indexes, by exploiting the self-similarity of images. However, the latency time of SwinIR-light is more than $\times10$ slower than the CNN-based methods, because the calculation of self-attention in SwinIR is a heavy burden for inference. Benefiting from our efficient long-range attention design, our ELAN-light model not only obtains the best or second best PSNR/SSIM indexes on all the five datasets and on all the three zooming scales, but also is about $\times4.5$ faster than SwinIR-light. Its \#Params and \#FLOPs are also smaller compared with SwinIR-light.


\textbf{Qualitative comparison}. We then qualitatively compare the SR quality of different light-weight models. The $\times4$ SR results on six example images are shown in Figure \ref{VISUAL_ELAN_LIGHT_figure} and Figure \ref{VISUAL_ELAN_LIGHT_figure_extra}. One can see that all the CNN-based models result in blurry and distorted edges on most images. The transformer based SwinIR-light can recover the main structure in the 1st image and partial edges in the 2nd image of Figure \ref{VISUAL_ELAN_LIGHT_figure}, but cannot correctly restore the structures among the other four images (e.g., ``Barbara", ``img004", ``img012" and ``BokuHaSitatakaKun"). Our ELAN-light is the only method that can recover the main structures on most of the images with clear and sharp edges, demonstrating the effectiveness of our modeling of long-range self-attention for the SR task. It is worth mentioning that the advantage of ELAN-light over SwinIR-light is achieved using much less computational cost, as we have validated in Table \ref{LightWeightBenchmarkResults}. 


\subsection{Comparison with classic SR models} 

To validate the scalability of ELAN, we further compare the normal version of ELAN with state-of-the-art 
classic performance-oriented SR models, including EDSR \cite{lim2017enhanced}, SRFBN \cite{li2019feedback}, RNAN \cite{zhang2019residual}, RDN \cite{zhang2018residual}, RCAN \cite{liu2020residual}, SAN \cite{dai2019second}, IGNN \cite{zhou2020cross}, HAN \cite{niu2020single}, NLSA \cite{mei2021image} and SwinIR \cite{liang2021swinir}. Note that these models employ either very deep network topology with large channel number or complicated self-attention and non-local strategies. 

\textbf{Quantitative comparison}. The quantitative results are shown in Table \ref{ClassicBenchmarkResults}. As can be seen, our ELAN achieves the best results on almost all benchmarks and all upscaling factors. In particular, compared with SwinIR, our ELAN obtains better PSNR and SSIM indexes on almost all settings with less number of parameters and FLOPs, and more than $\times2$ faster inference speed. These performance gains show that the proposed ELAB block succeeds in capturing useful information (such as long-range dependency) to conduct super-resolution in a more efficient and effective manner. Although the classical CNN-based SR models have huge amount of parameters and FLOPs, their performance is less competitive than the transformer-based models. Comparing with most CNN-based methods, our ELAN shows significant advantages in reconstruction performance metrics, benefiting from its larger receptive field and capability of modeling long-range spatial feature correlation. Some methods such as HAN and NLSA can also obtain competitive performance by exploiting dedicated attention mechanisms and very deep network, but their computation and memory cost are very expensive. For example, the NLSA model on $\times2$ upscaling task is too heavy to execute on a single NVIDIA 2080Ti GPU. Nevertheless, our ELAN obtains better performance than these complicated CNN-based models at much less cost.

\begin{table}[!t]
\caption{Performance comparison of different classic performance-oriented SR models on five benchmarks. PSNR/SSIM on Y channel are reported on each dataset. \#Params and \#FLOPs are the total number of network parameters and floating-point operations, respectively. Noted that all the efficiency proxies (\#Params, \#FLOPs and Latency) are measured under the setting of upscaling SR images to $1280\times720$ resolution on all scales. Best and second best PSNR/SSIM indexes are marked in \textcolor{red}{red} and \textcolor{blue}{blue} colors, respectively. The CNN-based methods and transformer-based methods are separated by a dash line. 'NaN' indicates that corresponding models are too heavy to run on a single NVIDIA 2080Ti GPU. '--' means that the result is not available.}
\label{ClassicBenchmarkResults}
\resizebox{\textwidth}{50mm}{
    \begin{tabular}{c|c|c|c|c|c|c|c|c|c}
    \hline
    Scale & Model & \makecell{\#Params \\ (K)}& \makecell{\#FLOPs \\ (G)} & \makecell{Latency\\(ms)} & \makecell{Set5 \cite{bevilacqua2012low}\\PSNR/SSIM} & \makecell{Set14 \cite{zeyde2010single}\\PSNR/SSIM} & \makecell{B100\cite{martin2001database}\\PSNR/SSIM} & \makecell{Urban100 \cite{huang2015single}\\PSNR/SSIM} & \makecell{Manga109 \cite{matsui2017sketch}\\PSNR/SSIM} \\
    \hline
    \multirow{11}{*}{$\times$ 2}
        & EDSR \cite{lim2017enhanced} & 40730 & 9387 & 1143 & 38.11/0.9602 & 33.92/0.9195 & 32.32/0.9013 & 32.93/0.9351 & 39.10/0.9773\\
        & SRFBN \cite{li2019feedback} & 2140 & 5044 & 920 & 38.11/0.9609 & 33.82/0.9196 & 32.29/0.9010 & 32.62/0.9328 & 39.08/0.9779\\ 
        & RNAN \cite{zhang2019residual} & 9107 & NaN & NaN & 38.17/0.9611 & 33.87/0.9207 & 32.32/0.9014 & 32.73/0.9340 & 39.23/0.9785\\
        & RDN \cite{zhang2018residual} & 22123 & 5098 & 846 & 38.24/0.9614 & 34.01/0.9212 & 32.34/0.9017 & 32.89/0.9353 & 39.18/0.9780\\ 
        & OISR \cite{he2019ode} & 41910 & 9657 & --- & 38.21/0.9612 & 33.94/0.9206 & 32.36/0.9019 & 33.03/0.9365 & --- \\
        & RCAN \cite{zhang2018residual} & 15445 & 3530 & 743 & 38.27/0.9614 & 34.12/0.9216 & 32.41/0.9027 & 33.34 0.9384 & 39.44/0.9786\\
        & SAN \cite{dai2019second} & 15861 & 3050 & NaN & 38.31/0.9620 & 34.07/0.9213 & 32.42/0.9028 & 33.10/0.9370 & 39.32/0.9792\\
        & IGNN \cite{zhou2020cross} & 49513 & --- & --- & 38.24/0.9613 & 34.07/0.9217 & 32.41/0.9025 & 33.23/0.9383 & 39.35/0.9786\\
        & HAN \cite{niu2020single} & 63608 & 14551 & 2278 & 38.27/0.9614 & \textcolor{blue}{34.16}/0.9217 & 32.41/0.9027 & 33.35/0.9385 & 39.46/0.9785\\
        & NLSA \cite{mei2021image} & 41796 & 9632 & NaN & 38.34/0.9618 & 34.08/\textcolor{red}{0.9231} & 32.43/\textcolor{blue}{0.9027} & 33.42/\textcolor{red}{0.9394} & 39.59/0.9789\\
        \cdashline{2-10}[5pt/5pt]
        & SwinIR \cite{liang2021swinir} & 11752 & 2301 & 2913 & \color{blue}{38.35/0.9620} & 34.14/0.9227 & \color{blue}{32.44}/\textcolor{red}{0.9030} & \color{blue}{33.40/0.9393} & \color{blue}{39.60/0.9792} \\
        & ELAN (ours) & 8254 & 1965 & 1244 & \color{red}{38.36/0.9620} & \color{red}{34.20}/\textcolor{blue}{0.9228} & \color{red}{32.45/0.9030} & \textcolor{red}{33.44}/0.9391 & \color{red}{39.62/0.9793} \\
    \hline
    \hline
    \multirow{11}{*}{$\times$ 3}
        & EDSR \cite{lim2017enhanced} & 43680 & 4470 & 573 & 34.65/0.9280 & 30.52/0.8462 & 29.25/0.8093 & 28.80/0.8653 & 34.17/0.9476\\
        & SRFBN \cite{li2019feedback} & 2833 & 6024 & 672 & 34.70/0.9292 & 30.51/0.8461 & 29.24/0.8084 & 28.73/0.8641 & 34.18/0.9481 \\
        & RNAN \cite{zhang2019residual} & 9292 & 809 & NaN & 34.66/0.9290 & 30.52/0.8462 & 29.26/0.8090 & 28.75/0.8646 & 34.25/0.9483\\
        & RDN \cite{zhang2018residual} & 22308 & 2282 & 406 & 34.71/0.9296 & 30.57/0.8468 & 29.26/0.8093 & 28.80/0.8653 & 34.13/0.9484\\
        & OISR \cite{he2019ode} & 44860 & 4590 & --- & 34.72/0.9297 & 30.57/0.8470 & 29.29/0.8103 & 28.95/0.8680 & --- \\
        & RCAN \cite{zhang2018residual} & 15629 & 1586 & 367 & 34.74/0.9299 & 30.65/0.8482 & 29.32/0.8111 & 29.09/0.8702 & 34.44/0.9499\\
        & SAN \cite{dai2019second} & 15897 & 1620 & NaN & 34.75/0.9300 & 30.59/0.8476 & 29.33/0.8112 & 28.93/0.8671 & 34.30/0.9494\\
        & IGNN \cite{zhou2020cross} & 49512 & --- & --- & 34.72/0.9298 & 30.66/0.8484 & 29.31/0.8105 & 29.03/0.8696 & 34.39/0.9496\\
        & HAN \cite{niu2020single} & 64346 & 6534 & 1014 & 34.75/0.9299 & 30.67/0.8483 & 29.32/0.8110 & 29.10/0.8705 & 34.48/0.9500\\
        & NLSA \cite{mei2021image} & 44747 & 4579 & 840 & 34.85/0.9306 & 30.70/0.8485 & 29.34/0.8117 & 29.25/0.8726 & 34.57 0.9508\\
        \cdashline{2-10}[5pt/5pt]
        & SwinIR \cite{liang2021swinir} & 11937 & 1026 & 1238 & \textcolor{blue}{34.89/0.9312} & \textcolor{blue}{30.77/0.8503} & \textcolor{blue}{29.37/0.8124} & \textcolor{blue}{29.29/0.8744} & \textcolor{red}{34.74/0.9518}\\
        & ELAN (ours) & 8278 & 874 & 530 & \textcolor{red}{34.90/0.9313} & \textcolor{red}{30.80/0.8504} & \textcolor{red}{29.38/0.8124} & \textcolor{red}{29.32/0.8745} & \textcolor{blue}{34.73/0.9517} \\
    \hline
    \hline
    \multirow{11}{*}{$\times$ 4}
        & EDSR \cite{lim2017enhanced} & 43090 & 2895 & 360 & 32.46/0.8968 & 28.80/0.7876 & 27.71/0.7420 & 26.64/0.8033 & 31.02/0.9148\\
        & SRFBN \cite{li2019feedback} & 3631 & 7466 & 551 & 32.47/0.8983 & 28.81/0.7868 & 27.72/0.7409 & 26.60/0.8015 & 31.15/0.9160\\
        & RNAN \cite{zhang2019residual} & 9255 & 480 & NaN & 32.49/0.8982 & 28.83/0.7878 & 27.72/0.7421 & 26.61/0.8023 & 31.09/0.9149\\
        & RDN \cite{zhang2018residual} & 22271 & 1310 & 243 & 32.47/0.8990 & 28.81/0.7871 & 27.72/0.7419 & 26.61/0.8028 & 31.00 0.9151\\
        & OISR \cite{he2019ode} & 44270 & 2963 & --- & 32.53/0.8992 & 28.86/0.7878 & 27.75/0.7428 & 26.79/0.8068 & --- \\
        & RCAN \cite{zhang2018residual} & 15592 & 918 & 223 & 32.63/0.9002 & 28.87/0.7889 & 27.77/0.7436 & 26.82/0.8087 & 31.22/0.9173\\
        & SAN \cite{dai2019second} & 15861 & 937 & NaN & 32.64/0.9003 & 28.92/0.7888 & 27.78/0.7436 & 26.79/0.8068 & 31.18/0.9169\\
        & IGNN \cite{zhou2020cross} & 49513 & --- & --- & 32.57/0.8998 & 28.85/0.7891 & 27.77/0.7434 & 26.84/0.8090 & 31.28 0.9182\\
        & HAN \cite{niu2020single} & 64199 & 3776 & 628 & 32.64/0.9002 & 28.90/0.7890 & 27.80/0.7442 & 26.85/0.8094 & 31.42/0.9177\\
        & NLSA \cite{mei2021image} & 44157 & 2956 & 502 & 32.59/0.9000 & 28.87/0.7891 & 27.78/0.7444 & 26.96/0.8109 & 31.27/0.9184\\
        \cdashline{2-10}[5pt/5pt]
        & SwinIR \cite{liang2021swinir} & 11900 & 584 & 645 & \textcolor{blue}{32.72/0.9021} & \textcolor{blue}{28.94/0.7914} & \textcolor{blue}{27.83/0.7459} & \textcolor{blue}{27.07/0.8164} & \textcolor{blue}{31.67/0.9226}\\
        & ELAN (ours) & 8312 & 494 & 298 & \textcolor{red}{32.75/0.9022} & \textcolor{red}{28.96/0.7914} & \textcolor{red}{27.83/0.7459} & \textcolor{red}{27.13/0.8167} & \textcolor{red}{31.68/0.9226} \\
    \hline
    \end{tabular}
}
\end{table}

\begin{figure}[!t]
  \centering
  \includegraphics[width=1.0\linewidth]{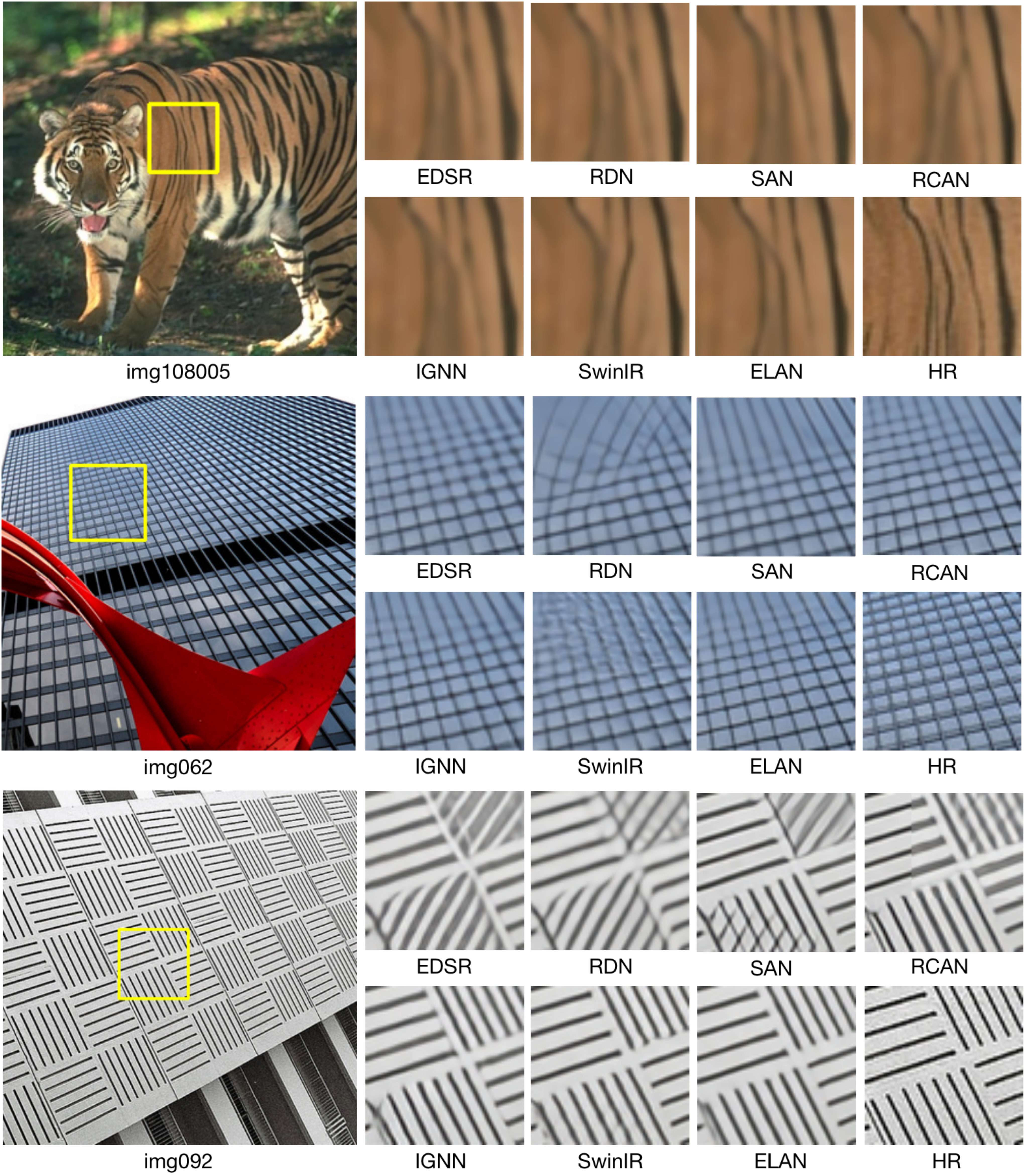}
  \caption{Qualitative comparison of state-of-the-art classic SR models for $\times4$ upscaling task. The ELAN can restore more accurate and sharper details than the other models.}
  \label{VISUAL_ELAN_CLASSIC_figure}
\end{figure}

\begin{figure}[!t]
  \centering
  \includegraphics[width=1.0\linewidth]{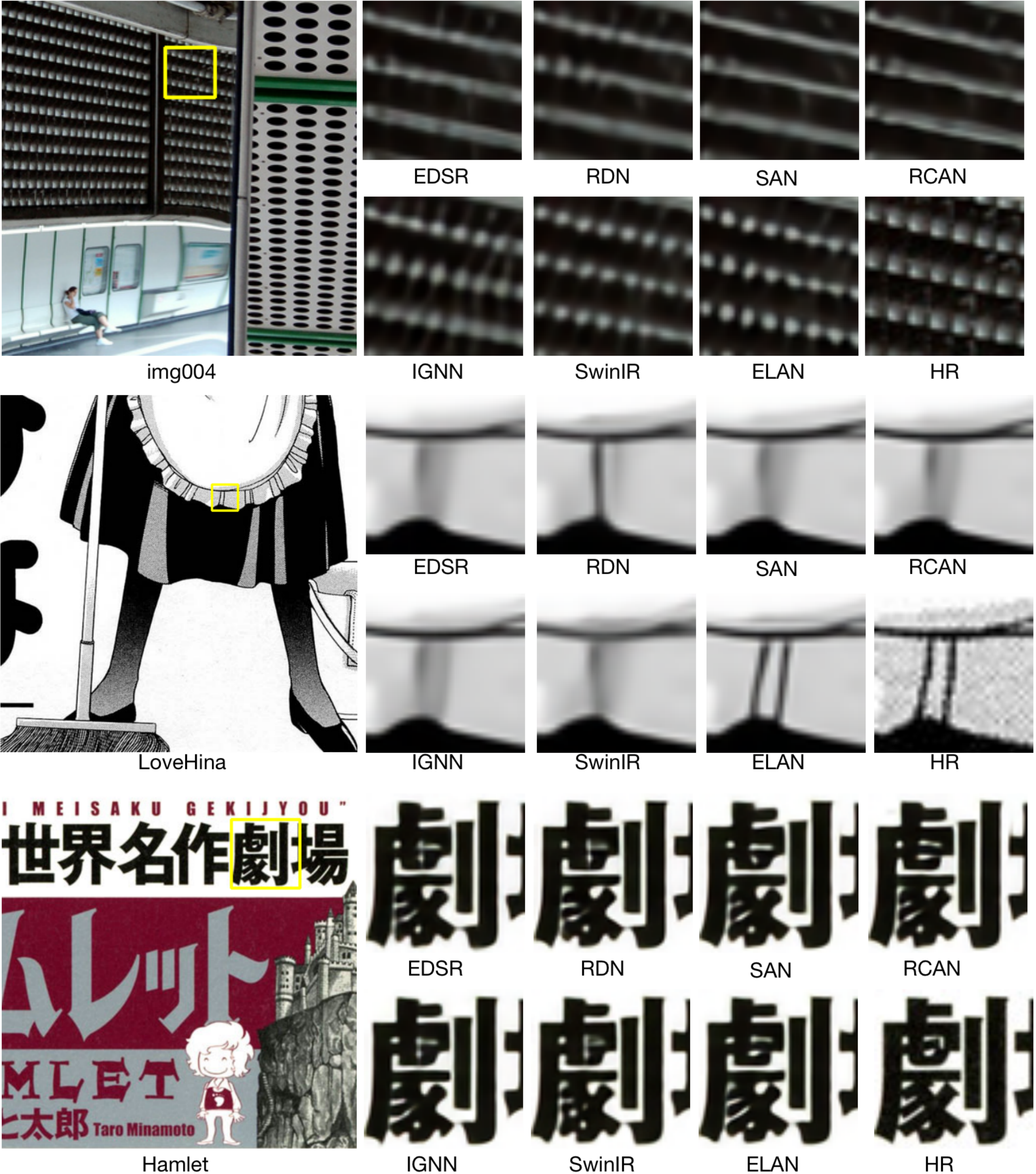}
  \caption{Qualitative comparison of state-of-the-art classic SR models for $\times4$ upscaling task. The ELAN can restore more accurate and sharper details than the other models.}
  \label{VISUAL_ELAN_CLASSIC_figure_extra}
\end{figure}

\textbf{Qualitative comparison}. We compare the visual quality of SR results by our ELAN and six representative models, including EDSR \cite{lim2017enhanced}, RDN \cite{zhang2018residual}, SAN \cite{dai2019second}, RCAN \cite{zhang2018residual}, IGNN \cite{li2019feedback} and SwinIR \cite{liang2021swinir}, on $\times 4$ upscaling task. The results on six example images are shown in Figure \ref{VISUAL_ELAN_CLASSIC_figure} and Figure \ref{VISUAL_ELAN_CLASSIC_figure_extra}. 
One can first observe that most of the compared methods yield either blurry or inaccurate edges and textures (see the fur texture of the tiger body in Figure \ref{VISUAL_ELAN_CLASSIC_figure} and the edges on the skirt in Figure \ref{VISUAL_ELAN_CLASSIC_figure_extra}), while our ELAN can restore more accurate and sharper edges. The advantage of ELAN is more obvious on the images with repetitive patterns and edges (see the 2nd and 3rd images of of Figure \ref{VISUAL_ELAN_CLASSIC_figure} and 1st image of Figure \ref{VISUAL_ELAN_CLASSIC_figure_extra}). Most of the compared methods fail to reconstruct the correct structures, and even generate undesired artifacts. The transformer based methods such as SwinIR \cite{liang2021swinir} can reproduce most of the patterns and edges in ``img092" and ``img003" but also suffer from artifacts and distortions in ``img062" and ``img092". In contrast, our ELAN can recover very clear patterns and edges in these examples. Referring to the last example in Figure \ref{VISUAL_ELAN_CLASSIC_figure_extra} where the Chinese character consists of intensive strokes of many directions, our ELAN is the only model that can reproduce distinct edges among all strokes. These comparisons clearly demonstrate the advantage of ELAN in recovering image structures and details from low-resolution inputs. 

\subsection{Ablation studies}
To better understand how ELAN works, we present comprehensive ablation studies to evaluate the roles of different components of ELAN, as well as the depth selection for the shared attention mechanism.

\textbf{The components of ELAN.} We first utilize the architecture and setting of ELAN-light to ablate the roles of different components of ELAB, and observe the change in performance and efficiency. The PSNR/SSIM indexes on five benchmarks and inference time of variant models are reported in Table \ref{ablation_elan_design}. The indexes of SwinIR-light is also listed for reference. Specifically, we start with a naive baseline by removing the redundant sub-branches from SwinIR-light, where the deep feature extraction module is composed of 24 sequential swin transformer block. As expected, this simplification leads to slight performance drop compared with the SwinIR-light while reducing the latency from 271ms to 247ms. By utilizing our shifted windows mechanism, the performance remain almost unchanged while the inference speed is decreased from 247ms to 177ms. 

We then replace the SA calculation with our proposed ASA, the inference latency is greatly reduced from 177ms to 66ms without losing PSNR/SSIM performance. By employing the GMSA which can efficiently model long-range dependency, the PSNR and SSIM indexes get significantly improved on all five datasets. Specifically, the PSNR increases by 0.21dB and 0.19dB on the Urban100 and Manga109 datasets, respectively, while the latency is only slightly increased by 9ms. This indicates the effectiveness of GMSA over the SA in SwinIR with small fixed window size. Finally, by employing the proposed shared attention mechanism, we can further speed up the inference time of ELAN-light with little performance drop. Combining all the improvements, the final version of ELAN-light achieves about $\times4.5$ acceleration while maintaining comparable performance to SwinIR-light.

\begin{table}[t]
\caption{Ablation study on network design for ELAN.}
\label{ablation_elan_design}
\resizebox{\textwidth}{14mm}{
    \begin{tabular}{c|c|c|c|c|c|c|c|c|c|c|c|c|c}
    \hline
    
    \multirow{3}{*}{Scale} &
    \multirow{3}{*}{Model} & 
    \multicolumn{4}{c|}{Different components} &
    \multirow{3}{*}{\makecell{\#Params\\(K)}} & 
    \multirow{3}{*}{\makecell{\#FLOPs\\(G)}} &
    \multirow{3}{*}{Latency(ms)} &
    Set5 \cite{bevilacqua2012low} & 
    Set14 \cite{zeyde2010single} & 
    B100\cite{martin2001database} & 
    U100 \cite{huang2015single} & 
    Manga109 \cite{matsui2017sketch}\\
    \cline{3-6}
    \cline{10-14}
    & & \makecell{Shifted \\ window} & ASA & GMSA & \makecell{Shared\\ attention} & & & & PSNR/SSIM & PSNR/SSIM & PSNR/SSIM & PSNR/SSIM & PSNR/SSIM\\
    \hline
    \multirow{5}{*}{$\times$ 4}
        & SwinIR-light \cite{liang2021swinir} &  &  &  &  & 897 & 49.6 & 271 & 32.44/0.8976 & 28.77/0.7858 & 27.69/0.7406 & 26.47/0.7980 & 30.92/0.9151 \\
        & ELAN-light &  &  &  &  & 767 & 44.6 & 247 & 32.38/0.8971 & 28.68/0.7832 & 27.62/0.7368 & 26.40/0.7973 & 30.78/0.9142\\
        & ELAN-light & \checkmark &  &  &  & 765 & 44.4 & 177 & 32.38/0.8971 & 28.69/0.7833 & 27.61/0.7367 & 26.42/0.7973 & 30.78/0.9141 \\
        & ELAN-light & \checkmark & \checkmark &  &  & 641 & 43.8 & 66 & 32.39/0.8970 & 28.67/0.7831 & 27.62/0.7368 & 26.39/0.7972 & 30.76/0.9139 \\
        & ELAN-light & \checkmark & \checkmark & \checkmark &  & 641 & 45.5 & 75 &32.47/0.8977 & 28.79/0.7858 & 27.71/0.7405 & 26.60/0.7985 & 30.95/0.9151 \\
        & ELAN-light & \checkmark & \checkmark & \checkmark & \checkmark & 601 & 43.2 & 62 & 32.43/0.8975 & 28.78/0.7858 & 27.69/0.7406 & 26.54/0.7982 & 30.92/0.9150 \\
    \hline
    \hline
    \end{tabular}
}
\end{table}



\textbf{Depth of shared attention}. We further conduct a detailed ablation study on the depth of shared attention blocks (i.e., $n$ in Figure \ref{ELAB_mechanism_figure}(b)) in sequential layers. ELAN-light is employed for this study and the depth of shared attention blocks varies in \{0, 1, 3, 5\}. Note that $n=0$ means that each ELAB calculates self-attention independently. The results of ELAN-light using different $n$ are reported in Table \ref{ablation_elan_shared_mechanism}. We also provide one representative CNN-based method EDSR-baseline for intuitive comparison. One can observe that using larger $n$ can effectively reduce the number of network parameters, FLOPs and latency time in the inference stage, at the cost of performance drop.
By choosing an appropriate $n$, our ELAN can achieve a good trade-off between efficiency and performance. It is worth mentioning that even using $n=5$, the PSNR of ELAN-light on the challenging Urban100 and Manga109 still outperform EDSR-baseline by a large margin (up to 0.31 dB and 0.28 dB, respectively), validating the advantage of modeling long-range attention in our model.




\begin{table}[t]
\caption{Ablation study on depth selection of shared attention.}
\resizebox{\textwidth}{12mm}{
    \begin{tabular}{c|c|c|c|c|c|c|c|c|c|c}
    \hline
    Scale & Model & Depth($n$) & \makecell{\#Params\\(K)} & \makecell{\#FLOPs\\(G)} & \makecell{Latency (ms)} & \makecell{Set5 \cite{bevilacqua2012low}\\PSNR/SSIM} & \makecell{Set14 \cite{zeyde2010single}\\PSNR/SSIM} & \makecell{B100\cite{martin2001database}\\PSNR/SSIM} & \makecell{U100 \cite{huang2015single}\\PSNR/SSIM} & \makecell{Manga109 \cite{matsui2017sketch}\\PSNR/SSIM}\\
    \hline
    \multirow{4}{*}{$\times$ 4}
        & ELAN-light & 0 & 641 & 45.5 & 75 & 32.47/0.8977 & 28.79/0.7858 & 27.71/0.7405 & 26.60/0.7985 & 30.95/0.9151\\
        & ELAN-light & 1 & 601 & 43.2 & 62 & 32.43/0.8976 & 28.78/0.7858 & 27.69/0.7406 & 26.54/0.7982 & 30.92/0.9150\\
        & ELAN-light & 3 & 575 & 40.4 & 51 & 32.35/0.8970 & 28.71/0.7852 & 27.65/0.7401 & 26.40/0.7971 & 30.74/0.9133 \\
        & ELAN-light & 5 & 571 & 39.3 & 48 & 32.31/0.8967 & 28.66/0.7846 & 27.62/0.7400 & 26.35/0.7967 & 30.63/0.9120 \\
        \hline
        & EDSR-baseline \cite{lim2017enhanced} & --- & 1518 & 114.0 & 28 & 32.09/0.8938 & 28.58/0.7813 & 27.57/0.7357 & 26.04/0.7849 & 30.35/0.9067 \\
    \hline
    \end{tabular}
}
\label{ablation_elan_shared_mechanism}

\end{table}

\section{Conclusion}

In this paper, we proposed an efficient long-range attention network (ELAN) for single image super resolution. The ELAN had a neat topology with sequentially cascaded efficient long-range attention blocks (ELAB). Each ELAB was composed of a local feature extraction module with two sequential shift-conv and a group-wise multi-scale self-attention (GMSA) module to gradually increase the receptive field of self-attention (SA). Benefiting from our accelerated SA calculation and shared attention mechanism, ELAB can effectively capture the local structure and long-range dependency in a very efficient manner.
Extensive experiments show that ELAN can obtain highly competitive performance than previous state-of-the-art SR models on both light-weight and  performance-oriented settings, while being much more economical than previous transformer-based SR methods. Although our ELAN achieves significant speedup than SwinIR, the calculation of SA is still computation- and memory-intensive compared to those light-weight CNN-based models. In the future, we will further explore more efficient implementations or approximations of SA for more low-level vision tasks.  



%
%
\bibliographystyle{splncs04}
\bibliography{main}

\begin{thebibliography}{10}
\providecommand{\url}[1]{\texttt{#1}}
\providecommand{\urlprefix}{URL }
\providecommand{\doi}[1]{https://doi.org/#1}

\bibitem{agustsson2017ntire}
Agustsson, E., Timofte, R.: Ntire 2017 challenge on single image
  super-resolution: Dataset and study. In: Proceedings of the IEEE conference
  on computer vision and pattern recognition workshops. pp. 126--135 (2017)

\bibitem{ahn2018fast}
Ahn, N., Kang, B., Sohn, K.A.: Fast, accurate, and lightweight super-resolution
  with cascading residual network. In: Proceedings of the European Conference
  on Computer Vision (ECCV). pp. 252--268 (2018)

\bibitem{bevilacqua2012low}
Bevilacqua, M., Roumy, A., Guillemot, C., Alberi-Morel, M.L.: Low-complexity
  single-image super-resolution based on nonnegative neighbor embedding  (2012)

\bibitem{brown2020language}
Brown, T., Mann, B., Ryder, N., Subbiah, M., Kaplan, J.D., Dhariwal, P.,
  Neelakantan, A., Shyam, P., Sastry, G., Askell, A., et~al.: Language models
  are few-shot learners. Advances in neural information processing systems
  \textbf{33},  1877--1901 (2020)

\bibitem{buades2005non}
Buades, A., Coll, B., Morel, J.M.: A non-local algorithm for image denoising.
  In: 2005 IEEE Computer Society Conference on Computer Vision and Pattern
  Recognition (CVPR'05). vol.~2, pp. 60--65. IEEE (2005)

\bibitem{caballero2017real}
Caballero, J., Ledig, C., Aitken, A., Acosta, A., Totz, J., Wang, Z., Shi, W.:
  Real-time video super-resolution with spatio-temporal networks and motion
  compensation. In: Proceedings of the IEEE Conference on Computer Vision and
  Pattern Recognition. pp. 4778--4787 (2017)

\bibitem{cao2021swin}
Cao, H., Wang, Y., Chen, J., Jiang, D., Zhang, X., Tian, Q., Wang, M.:
  Swin-unet: Unet-like pure transformer for medical image segmentation. arXiv
  preprint arXiv:2105.05537  (2021)

\bibitem{carion2020end}
Carion, N., Massa, F., Synnaeve, G., Usunier, N., Kirillov, A., Zagoruyko, S.:
  End-to-end object detection with transformers. In: European conference on
  computer vision. pp. 213--229. Springer (2020)

\bibitem{chen2021pre}
Chen, H., Wang, Y., Guo, T., Xu, C., Deng, Y., Liu, Z., Ma, S., Xu, C., Xu, C.,
  Gao, W.: Pre-trained image processing transformer. In: Proceedings of the
  IEEE/CVF Conference on Computer Vision and Pattern Recognition. pp.
  12299--12310 (2021)

\bibitem{dai2019second}
Dai, T., Cai, J., Zhang, Y., Xia, S.T., Zhang, L.: Second-order attention
  network for single image super-resolution. In: Proceedings of the IEEE/CVF
  conference on computer vision and pattern recognition. pp. 11065--11074
  (2019)

\bibitem{dong2015image}
Dong, C., Loy, C.C., He, K., Tang, X.: Image super-resolution using deep
  convolutional networks. IEEE transactions on pattern analysis and machine
  intelligence  \textbf{38}(2),  295--307 (2015)

\bibitem{dong2016accelerating}
Dong, C., Loy, C.C., Tang, X.: Accelerating the super-resolution convolutional
  neural network. In: European conference on computer vision. pp. 391--407.
  Springer (2016)

\bibitem{dosovitskiy2020image}
Dosovitskiy, A., Beyer, L., Kolesnikov, A., Weissenborn, D., Zhai, X.,
  Unterthiner, T., Dehghani, M., Minderer, M., Heigold, G., Gelly, S., et~al.:
  An image is worth 16x16 words: Transformers for image recognition at scale.
  arXiv preprint arXiv:2010.11929  (2020)

\bibitem{fedus2021switch}
Fedus, W., Zoph, B., Shazeer, N.: Switch transformers: Scaling to trillion
  parameter models with simple and efficient sparsity. arXiv preprint
  arXiv:2101.03961  (2021)

\bibitem{he2019ode}
He, X., Mo, Z., Wang, P., Liu, Y., Yang, M., Cheng, J.: Ode-inspired network
  design for single image super-resolution. In: Proceedings of the IEEE/CVF
  Conference on Computer Vision and Pattern Recognition. pp. 1732--1741 (2019)

\bibitem{huang2017densely}
Huang, G., Liu, Z., Van Der~Maaten, L., Weinberger, K.Q.: Densely connected
  convolutional networks. In: Proceedings of the IEEE conference on computer
  vision and pattern recognition. pp. 4700--4708 (2017)

\bibitem{huang2015single}
Huang, J.B., Singh, A., Ahuja, N.: Single image super-resolution from
  transformed self-exemplars. In: Proceedings of the IEEE conference on
  computer vision and pattern recognition. pp. 5197--5206 (2015)

\bibitem{hui2019lightweight}
Hui, Z., Gao, X., Yang, Y., Wang, X.: Lightweight image super-resolution with
  information multi-distillation network. In: Proceedings of the 27th ACM
  International Conference on Multimedia. pp. 2024--2032 (2019)

\bibitem{ioffe2015batch}
Ioffe, S., Szegedy, C.: Batch normalization: Accelerating deep network training
  by reducing internal covariate shift. In: International conference on machine
  learning. pp. 448--456. PMLR (2015)

\bibitem{kim2016accurate}
Kim, J., Kwon~Lee, J., Mu~Lee, K.: Accurate image super-resolution using very
  deep convolutional networks. In: Proceedings of the IEEE conference on
  computer vision and pattern recognition. pp. 1646--1654 (2016)

\bibitem{kingma2014adam}
Kingma, D.P., Ba, J.: Adam: A method for stochastic optimization. arXiv
  preprint arXiv:1412.6980  (2014)

\bibitem{lai2017deep}
Lai, W.S., Huang, J.B., Ahuja, N., Yang, M.H.: Deep laplacian pyramid networks
  for fast and accurate super-resolution. In: Proceedings of the IEEE
  conference on computer vision and pattern recognition. pp. 624--632 (2017)

\bibitem{lecun2015deep}
LeCun, Y., Bengio, Y., Hinton, G.: Deep learning. nature  \textbf{521}(7553),
  436--444 (2015)

\bibitem{ledig2017photo}
Ledig, C., Theis, L., Husz{\'a}r, F., Caballero, J., Cunningham, A., Acosta,
  A., Aitken, A., Tejani, A., Totz, J., Wang, Z., et~al.: Photo-realistic
  single image super-resolution using a generative adversarial network. In:
  Proceedings of the IEEE conference on computer vision and pattern
  recognition. pp. 4681--4690 (2017)

\bibitem{li2020lapar}
Li, W., Zhou, K., Qi, L., Jiang, N., Lu, J., Jia, J.: Lapar: Linearly-assembled
  pixel-adaptive regression network for single image super-resolution and
  beyond. Advances in Neural Information Processing Systems  \textbf{33},
  20343--20355 (2020)

\bibitem{li2021localvit}
Li, Y., Zhang, K., Cao, J., Timofte, R., Van~Gool, L.: Localvit: Bringing
  locality to vision transformers. arXiv preprint arXiv:2104.05707  (2021)

\bibitem{li2019feedback}
Li, Z., Yang, J., Liu, Z., Yang, X., Jeon, G., Wu, W.: Feedback network for
  image super-resolution. In: Proceedings of the IEEE/CVF Conference on
  Computer Vision and Pattern Recognition. pp. 3867--3876 (2019)

\bibitem{liang2021swinir}
Liang, J., Cao, J., Sun, G., Zhang, K., Van~Gool, L., Timofte, R.: Swinir:
  Image restoration using swin transformer. In: Proceedings of the IEEE/CVF
  International Conference on Computer Vision. pp. 1833--1844 (2021)

\bibitem{lim2017enhanced}
Lim, B., Son, S., Kim, H., Nah, S., Mu~Lee, K.: Enhanced deep residual networks
  for single image super-resolution. In: Proceedings of the IEEE conference on
  computer vision and pattern recognition workshops. pp. 136--144 (2017)

\bibitem{liu2018non}
Liu, D., Wen, B., Fan, Y., Loy, C.C., Huang, T.S.: Non-local recurrent network
  for image restoration. Advances in neural information processing systems
  \textbf{31} (2018)

\bibitem{liu2020residual}
Liu, J., Tang, J., Wu, G.: Residual feature distillation network for
  lightweight image super-resolution. arXiv preprint arXiv:2009.11551  (2020)

\bibitem{liu2020deep}
Liu, L., Ouyang, W., Wang, X., Fieguth, P., Chen, J., Liu, X., Pietik{\"a}inen,
  M.: Deep learning for generic object detection: A survey. International
  journal of computer vision  \textbf{128}(2),  261--318 (2020)

\bibitem{liu2019roberta}
Liu, Y., Ott, M., Goyal, N., Du, J., Joshi, M., Chen, D., Levy, O., Lewis, M.,
  Zettlemoyer, L., Stoyanov, V.: Roberta: A robustly optimized bert pretraining
  approach. arXiv preprint arXiv:1907.11692  (2019)

\bibitem{liu2021transformer}
Liu, Y., Sun, G., Qiu, Y., Zhang, L., Chhatkuli, A., Van~Gool, L.: Transformer
  in convolutional neural networks. arXiv preprint arXiv:2106.03180  (2021)

\bibitem{liu2021swin}
Liu, Z., Lin, Y., Cao, Y., Hu, H., Wei, Y., Zhang, Z., Lin, S., Guo, B.: Swin
  transformer: Hierarchical vision transformer using shifted windows. In:
  Proceedings of the IEEE/CVF International Conference on Computer Vision. pp.
  10012--10022 (2021)

\bibitem{lu2021efficient}
Lu, Z., Liu, H., Li, J., Zhang, L.: Efficient transformer for single image
  super-resolution. arXiv preprint arXiv:2108.11084  (2021)

\bibitem{luo2020latticenet}
Luo, X., Xie, Y., Zhang, Y., Qu, Y., Li, C., Fu, Y.: Latticenet: Towards
  lightweight image super-resolution with lattice block. In: European
  Conference on Computer Vision. pp. 272--289. Springer (2020)

\bibitem{martin2001database}
Martin, D., Fowlkes, C., Tal, D., Malik, J.: A database of human segmented
  natural images and its application to evaluating segmentation algorithms and
  measuring ecological statistics. In: Proceedings Eighth IEEE International
  Conference on Computer Vision. ICCV 2001. vol.~2, pp. 416--423. IEEE (2001)

\bibitem{matsui2017sketch}
Matsui, Y., Ito, K., Aramaki, Y., Fujimoto, A., Ogawa, T., Yamasaki, T.,
  Aizawa, K.: Sketch-based manga retrieval using manga109 dataset. Multimedia
  Tools and Applications  \textbf{76}(20),  21811--21838 (2017)

\bibitem{mei2021image}
Mei, Y., Fan, Y., Zhou, Y.: Image super-resolution with non-local sparse
  attention. In: Proceedings of the IEEE/CVF Conference on Computer Vision and
  Pattern Recognition. pp. 3517--3526 (2021)

\bibitem{niu2020single}
Niu, B., Wen, W., Ren, W., Zhang, X., Yang, L., Wang, S., Zhang, K., Cao, X.,
  Shen, H.: Single image super-resolution via a holistic attention network. In:
  European conference on computer vision. pp. 191--207. Springer (2020)

\bibitem{paszke2019pytorch}
Paszke, A., Gross, S., Massa, F., Lerer, A., Bradbury, J., Chanan, G., Killeen,
  T., Lin, Z., Gimelshein, N., Antiga, L., et~al.: Pytorch: An imperative
  style, high-performance deep learning library. Advances in neural information
  processing systems  \textbf{32},  8026--8037 (2019)

\bibitem{radford2018improving}
Radford, A., Narasimhan, K., Salimans, T., Sutskever, I.: Improving language
  understanding by generative pre-training  (2018)

\bibitem{ramachandran2019stand}
Ramachandran, P., Parmar, N., Vaswani, A., Bello, I., Levskaya, A., Shlens, J.:
  Stand-alone self-attention in vision models. Advances in Neural Information
  Processing Systems  \textbf{32} (2019)

\bibitem{sajjadi2017enhancenet}
Sajjadi, M.S., Scholkopf, B., Hirsch, M.: Enhancenet: Single image
  super-resolution through automated texture synthesis. In: Proceedings of the
  IEEE international conference on computer vision. pp. 4491--4500 (2017)

\bibitem{sajjadi2018frame}
Sajjadi, M.S., Vemulapalli, R., Brown, M.: Frame-recurrent video
  super-resolution. In: Proceedings of the IEEE Conference on Computer Vision
  and Pattern Recognition. pp. 6626--6634 (2018)

\bibitem{shi2016real}
Shi, W., Caballero, J., Husz{\'a}r, F., Totz, J., Aitken, A.P., Bishop, R.,
  Rueckert, D., Wang, Z.: Real-time single image and video super-resolution
  using an efficient sub-pixel convolutional neural network. In: Proceedings of
  the IEEE conference on computer vision and pattern recognition. pp.
  1874--1883 (2016)

\bibitem{tai2017image}
Tai, Y., Yang, J., Liu, X.: Image super-resolution via deep recursive residual
  network. In: Proceedings of the IEEE conference on computer vision and
  pattern recognition. pp. 3147--3155 (2017)

\bibitem{tai2017memnet}
Tai, Y., Yang, J., Liu, X., Xu, C.: Memnet: A persistent memory network for
  image restoration. In: Proceedings of the IEEE international conference on
  computer vision. pp. 4539--4547 (2017)

\bibitem{tao2017detail}
Tao, X., Gao, H., Liao, R., Wang, J., Jia, J.: Detail-revealing deep video
  super-resolution. In: Proceedings of the IEEE International Conference on
  Computer Vision. pp. 4472--4480 (2017)

\bibitem{timofte2017ntire}
Timofte, R., Agustsson, E., Van~Gool, L., Yang, M.H., Zhang, L.: Ntire 2017
  challenge on single image super-resolution: Methods and results. In:
  Proceedings of the IEEE conference on computer vision and pattern recognition
  workshops. pp. 114--125 (2017)

\bibitem{touvron2021training}
Touvron, H., Cord, M., Douze, M., Massa, F., Sablayrolles, A., J{\'e}gou, H.:
  Training data-efficient image transformers \& distillation through attention.
  In: International Conference on Machine Learning. pp. 10347--10357. PMLR
  (2021)

\bibitem{wang2019edvr}
Wang, X., Chan, K.C., Yu, K., Dong, C., Change~Loy, C.: Edvr: Video restoration
  with enhanced deformable convolutional networks. In: Proceedings of the
  IEEE/CVF Conference on Computer Vision and Pattern Recognition Workshops.
  pp.~0--0 (2019)

\bibitem{wang2018esrgan}
Wang, X., Yu, K., Wu, S., Gu, J., Liu, Y., Dong, C., Qiao, Y., Change~Loy, C.:
  Esrgan: Enhanced super-resolution generative adversarial networks. In:
  Proceedings of the European Conference on Computer Vision (ECCV) Workshops.
  pp.~0--0 (2018)

\bibitem{wang2021uformer}
Wang, Z., Cun, X., Bao, J., Liu, J.: Uformer: A general u-shaped transformer
  for image restoration. arXiv preprint arXiv:2106.03106  (2021)

\bibitem{wu2018shift}
Wu, B., Wan, A., Yue, X., Jin, P., Zhao, S., Golmant, N., Gholaminejad, A.,
  Gonzalez, J., Keutzer, K.: Shift: A zero flop, zero parameter alternative to
  spatial convolutions. In: Proceedings of the IEEE Conference on Computer
  Vision and Pattern Recognition. pp. 9127--9135 (2018)

\bibitem{wu2020visual}
Wu, B., Xu, C., Dai, X., Wan, A., Zhang, P., Yan, Z., Tomizuka, M., Gonzalez,
  J., Keutzer, K., Vajda, P.: Visual transformers: Token-based image
  representation and processing for computer vision. arXiv preprint
  arXiv:2006.03677  (2020)

\bibitem{zamir2021restormer}
Zamir, S.W., Arora, A., Khan, S., Hayat, M., Khan, F.S., Yang, M.H.: Restormer:
  Efficient transformer for high-resolution image restoration. arXiv preprint
  arXiv:2111.09881  (2021)

\bibitem{zeyde2010single}
Zeyde, R., Elad, M., Protter, M.: On single image scale-up using
  sparse-representations. In: International conference on curves and surfaces.
  pp. 711--730. Springer (2010)

\bibitem{zhang2018image}
Zhang, Y., Li, K., Li, K., Wang, L., Zhong, B., Fu, Y.: Image super-resolution
  using very deep residual channel attention networks. In: Proceedings of the
  European conference on computer vision (ECCV). pp. 286--301 (2018)

\bibitem{zhang2019residual}
Zhang, Y., Li, K., Li, K., Zhong, B., Fu, Y.: Residual non-local attention
  networks for image restoration. arXiv preprint arXiv:1903.10082  (2019)

\bibitem{zhang2018residual}
Zhang, Y., Tian, Y., Kong, Y., Zhong, B., Fu, Y.: Residual dense network for
  image super-resolution. In: Proceedings of the IEEE conference on computer
  vision and pattern recognition. pp. 2472--2481 (2018)

\bibitem{zheng2021rethinking}
Zheng, S., Lu, J., Zhao, H., Zhu, X., Luo, Z., Wang, Y., Fu, Y., Feng, J.,
  Xiang, T., Torr, P.H., et~al.: Rethinking semantic segmentation from a
  sequence-to-sequence perspective with transformers. In: Proceedings of the
  IEEE/CVF conference on computer vision and pattern recognition. pp.
  6881--6890 (2021)

\bibitem{zhou2020cross}
Zhou, S., Zhang, J., Zuo, W., Loy, C.C.: Cross-scale internal graph neural
  network for image super-resolution. Advances in neural information processing
  systems  \textbf{33},  3499--3509 (2020)

\end{thebibliography}
\end{document}